\documentclass[11pt,a4paper]{article}
\usepackage[hyperref]{eacl2021}
\usepackage{times}
\usepackage{latexsym}

\usepackage{booktabs}
\usepackage{multirow}
\usepackage{microtype}

\usepackage{graphicx}
\usepackage{multirow}
\usepackage{caption}
\usepackage{subcaption}
\usepackage{xcolor}
\usepackage{amssymb}

\aclfinalcopy 

\title{BERT-based Multi-Task Model for Country and Province Level Modern Standard Arabic and Dialectal Arabic Identification}

\author{Abdellah El Mekki$^1$ \hspace{0.5cm} Abdelkader El Mahdaouy$^1$ \hspace{0.5cm} Kabil Essefar$^1$\\ \textbf{Nabil El Mamoun$^2$ \hspace{0.5cm} Ismail Berrada$^1$ \hspace{0.5cm} \textbf{Ahmed Khoumsi}$^3$}\\
$^1$School of Computer Sciences, Mohammed VI Polytechnic University, Morocco \\
$^2$Faculty of Sciences Dhar EL Mahraz, Sidi Mohamed Ben Abdellah University, Morocco\\
$^3$Dept. Electrical \& Computer Engineering, University of Sherbrooke, Canada\\
{\tt \{firstname.lastname\}@um6p.ma} \\
{\tt ahmed.khoumsi@usherbrooke.ca}\\
}

\date{}

\begin{document}
\maketitle
\begin{abstract}
Dialect and standard language identification are crucial tasks for many Arabic natural language processing applications. In this paper, we present our deep learning-based system, submitted to the second NADI shared task for country-level and province-level identification of Modern Standard Arabic (MSA) and Dialectal Arabic (DA). The system is based on an end-to-end deep Multi-Task Learning (MTL) model to tackle both country-level and province-level MSA/DA identification.  The latter MTL model consists of a shared Bidirectional Encoder Representation Transformers (BERT) encoder, two task-specific attention layers, and two classifiers.  Our key idea is to leverage both the task-discriminative and the inter-task shared features for country and province MSA/DA identification.  
The obtained results show that our MTL model outperforms single-task models on most subtasks.
\end{abstract}

\begin{figure*}
 \centering
  \includegraphics[scale=0.3]{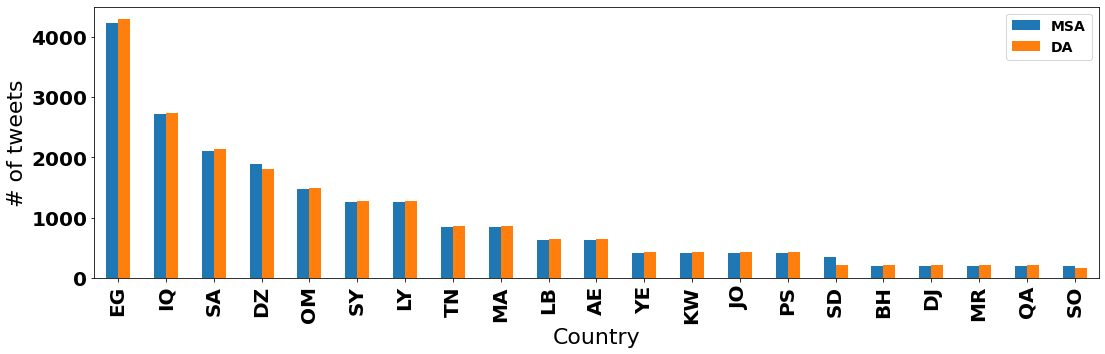}
  \caption{Label distribution of the training set for the country-level classification subtasks. Country code following the ISO 3166-1 alpha-2 (Wikipedia, 2020)}
  \label{fig:labels_distri}
\end{figure*}

\section{Introduction} 

The Arabic language is spoken by approximately 400 million people and characterized by different varieties. On the one hand, people of the Arab world tend to use Modern Standard Arabic (MSA) as a communication channel in formal situations (e.g. media, religion, education).  On the other hand, Arabic dialects are usually used for informal communication in daily life. These dialects differ, to varying degrees, from one region to another. 

Generally, existing research works categorized DA into four regions (Maghreb, Egypt, Gulf, and Levant)  based on the coarse-grained taxonomy \cite{zaidan_paper_taxonomy}. Recently, several research studies classified DA into more fine-grained varieties such as country-level dialects \cite{salameh-etal-2018-fine, bouamor:2019:madar, mageed:2020:nadi}.

In the last few years, Arabic dialect identification has gained much attention \cite{bouamor:2019:madar,el-mekki-etal-2020-weighted, mageed:2020:nadi,abdul-mageed-etal-2020-toward}. Identifying the dialect of an end-user is a very important task in many applications such as user profiling, personalized customer support, etc. Nevertheless, due to the nature and the structure of DA as well as MSA \cite{msa_morphological_complexity,habash2010introduction,habash-etal-2012-conventional}, this task faces several challenges. 

Previous works on DA identification mainly focused on the use of traditional machine learning models \cite{abu-kwaik-saad-2019-arbdialectid,meftouh-etal-2019-smart} and single-task deep learning models \cite{talafha-etal-2020-multi}. 
\cite{el-mekki-etal-2020-weighted} introduced hierarchical models that perform the training and prediction of the country-level and province-level classification based on a sequential process. Recently, \cite{abdul-mageed-etal-2020-toward} shown the effectiveness of MTL using MARBERT, a transformer-based language model pre-trained on a massive dataset of 128GB Arabic tweets, for both country-level and province-level DA identification. 

In this paper, we tackle both the DA and MSA identification of the second NADI (Nuanced Arabic Dialect Identification) shared tasks: Country-level MSA identification, country-level DA identification, province-level MSA identification, and province-level DA identification \cite{mageed:2021:nadi}. Our submitted system consists of  an end-to-end deep MLT model that predicts both the province and the country of a given Arabic tweet. Our model leverages MARBERT’s contextualized word embedding \cite{mageed2020marbert} with two task-specific attention layers that extract the task-discriminative features. The latter features are then concatenated with the encoder's pooled embedding ([CLS] embedding) and are fed to the task classifiers. Thus, the core idea of our approach is to combine both task-discriminative and inter-task shared features for country and province MSA/DA identification. The obtained results show that our MTL model outperforms its single-task counterpart on all evaluated subtasks. These results prove the effectiveness of combining task-specific features and inter-task shared features for country-level and province-level MSA and DA identification.

The rest of this paper is organized as follows. Section \ref{sec:data} describes the NADI shared task's datasets. Section \ref{sec:method} presents our proposed method. Section \ref{sec:results} summarizes the obtained results for both subtask 1 and subtask 2. In Section \ref{sec:discussion}, we discuss these results. Finally, the conclusion is given in section \ref{sec:conclusion}.

\section{Data} \label{sec:data}
The second NADI shared task consists of four subtasks on MSA and DA country-level as well as province-level identification. Table \ref{tab:subtasks} presents the four subtasks of NADI'2020.

\begin{table}[htbp]
  \centering
  \caption{NADI'2021 DA and MSA identification subtasks}
    \begin{tabular}{lcc}
\cmidrule{1-3}         
\textbf{Data} & \textbf{Country-level} & \textbf{Province-level} \\
\cmidrule{1-3}    \textbf{MSA} & Subtask 1.1 & Subtask 2.1 \\
    \midrule
    \textbf{DA} & Subtask 1.2 & Subtask 2.2 \\
    \bottomrule
    \end{tabular}%
  \label{tab:subtasks}%
\end{table}%

For both MSA and DA subtasks, the organizers of NADI’2021 provided a dataset of 31,000 labeled tweets covering 21 Arab countries and 100 Arab provinces. The training set consists of 21,000 tweets, while the rest 10,000 are equally distributed between the development and test sets. Finally, each tweet is assigned a single country label and a single province label. Figure \ref{fig:labels_distri} shows that the distribution of tweets for the country-level classification subtasks is unbalanced. Furthermore, Egypt, Iraq and Saudi Arabia countries have the highest number of tweets, while Mauritania, Qatar and Somalia have the lowest one.

\section{Method} \label{sec:method}

Our multi-task model, for both tasks, consists of three main components: BERT encoder (MARBERT pre-trained language model), two task-specific attention layers, and two task classifiers.

\subsection{BERT Encoder}
The input tweets are encoded using a Bidirectional Encoder Representation from Transformers (BERT) model \cite{devlin-etal-2019-bert}. BERT employs multiple transformer blocks to encode the input text. This model is trained on large textual corpora by jointly optimizing the Masked Language Model (MLM) and the Next Sentence Prediction (NSP) objectives. Fine-tuning BERT model on the downstream tasks has shown state-of-the-art performances in many NLP applications.


In order to avoid domain shift, our end-to-end model for NADI'2021 uses MARBERT. In fact, MARBERT \cite{mageed2020marbert} is a variation of BERT pre-trained on a large Arabic Twitter dataset (1 billion tweets) using only MLM objective (tweets are short). 


\subsection{Task-specific attention layer}

Two task-specific attention layers are used to reward tokens' hidden representation (contextual embedding) that contributes to the correct classification of tweets for the country-level and province-level tasks. These layers operate on top of the contextualized word embedding of the BERT encoder $H = [h_1, h_2, ...,h_n] \in \mathbb{R}^{n \times d}$, where $n$ is the sequence length and $d$ is the embedding dimension. Hence, each task-specific attention layer \cite{Bahdanau2015NeuralMT,yang-etal-2016-hierarchical} can attend to some parts of the tweet to extract the task-discriminative features. The attention mechanism is given by:
\[ C = tanh(H W_{a}) \]
\[ \alpha = softmax ( C^{T} W_{\alpha})\]
\[v = \alpha \cdot H^{T} \]
where $W_{a} \in \mathbb{R}^{d \times 1}$, $W_{\alpha} \in \mathbb{R}^{n \times n}$ are the attention mechanism's learnable parameters, $C \in \mathbb{R}^{n \times 1}$ and $\alpha \in [0,1]^{n}$ weights the word hidden representations according to their relevance to the task, $v$ represents the task-relevant information contained in a tweet. 
\subsection{Task classifier}
The task classifier consists of one hidden layer and one output layer. The pooled output ($h_{[CLS]}$ embedding) and the vector $v$, obtained using the task-specific attention layer, are concatenated and passed to the task classifier. The latter outputs the predicted task label. 

\subsection{Multi-task learning objective}
Our MTL model is trained to jointly optimize both tasks' cross-entropy losses. For the country-level MSA/DA identification, our model minimizes: 
\[
L_{Country}(\hat{y}^{c},y^{c}) = -\sum_{i=1}^{N}\sum_{j=1}^{l}y_{ij}^clog(\hat{y}^c_{ij})
\]

\noindent where $y_{ij}^{c}$ is the ground-truth label, $\hat{y}_{ij}^{c}$ is the predicted label, $N$ is  the number of training samples, and $l$ is the number of countries ($l = 21$).

\noindent For the province-level MSA/DA identification, our model is trained to minimize: 

\[L_{Province}(\hat{y}^{p},y^{p}) = -\sum_{i=1}^{N}\sum_{j=1}^{k}y^p_{ij}log(\hat{y}^p_{ij})\]
where $y_{ij}^{p}$ is the ground-truth label, $\hat{y}_{ij}^{p}$ is the predicted label, and $k$ is the number of provinces ($k= 100$).

\noindent Thus, the final loss of our model is:
\[L = L_{Country} + L_{Province}\]

Finally, our model is trained using Adam optimizer \cite{kingma2017adam}, with a learning rate of $1 \times 10^{-5}$. Based on several experiments, the batch size and the number of epochs are set to 16 and 5, respectively. For tweets cleaning, we have implemented the same preprocessing pipeline that is used by MARBERT which consists of diacritics removal and mention substitution by USER token.
\begin{table*}[!h]
\centering

\begin{tabular}{llc|c|c|c}
\cline{3-6}
                                                  &                                        & \multicolumn{2}{l|}{\textbf{Single-task model}} & \multicolumn{2}{l}{\textbf{Multi-task model}} \\ 
\cline{2-6}
\multicolumn{1}{l}{}                             & \multicolumn{1}{l|}{\textbf{Dev/Test}} & \textbf{F1}         & \textbf{Accuracy}         & \textbf{F1}          & \textbf{Accuracy}       \\ \cline{1-6}
\multicolumn{1}{l|}{\multirow{2}{*}{Subtask 1.1}} & \multicolumn{1}{l|}{Dev}               & 21.56               & 34.35                     & \textbf{23.05}       & \textbf{37.80}          \\ 
\multicolumn{1}{l|}{}                             & \multicolumn{1}{l|}{Test}              & 20.90                   & 33.73                         & \textbf{21.47}                    & \textbf{33.84}                       \\ 
\hline
\multicolumn{1}{l|}{\multirow{2}{*}{Subtask 1.2}} & \multicolumn{1}{l|}{Dev}               & 31.62               & 48.48                     & \textbf{32.04}       & \textbf{51.28}          \\ 
\multicolumn{1}{l|}{}                             & \multicolumn{1}{l|}{Test}              & 29.07                   & 49.50                         & \textbf{30.64}                    & \textbf{50.30}                       \\ 
\hline
\multicolumn{1}{l|}{\multirow{2}{*}{Subtask 2.1}} & \multicolumn{1}{l|}{Dev}               & 6.05                & 6.63                      & \textbf{6.40}        & \textbf{6.85}           \\ 
\multicolumn{1}{l|}{}                             & \multicolumn{1}{l|}{Test}              & 4.72                   & 5.00                         & \textbf{5.35}                    & \textbf{5.72}                       \\ \hline
\multicolumn{1}{l|}{\multirow{2}{*}{Subtask 2.2}} & \multicolumn{1}{l|}{Dev}               & 8.88                & 9.60                      & \textbf{9.40}        & \textbf{9.84}           \\ 
\multicolumn{1}{l|}{}                             & \multicolumn{1}{l|}{Test}              & 5.30                   & 6.90                         & \textbf{7.32}                    & \textbf{7.92}                      \\ 
\bottomrule
\end{tabular}

\caption{Scores of our models  (\%) for the 4 subtasks (dev-sets and test-sets).}
\label{table:results}
\end{table*}

\begin{figure*}[!h]
     \centering
     \begin{subfigure}[b]{0.49\textwidth}
         \centering
         \includegraphics[scale=0.130]{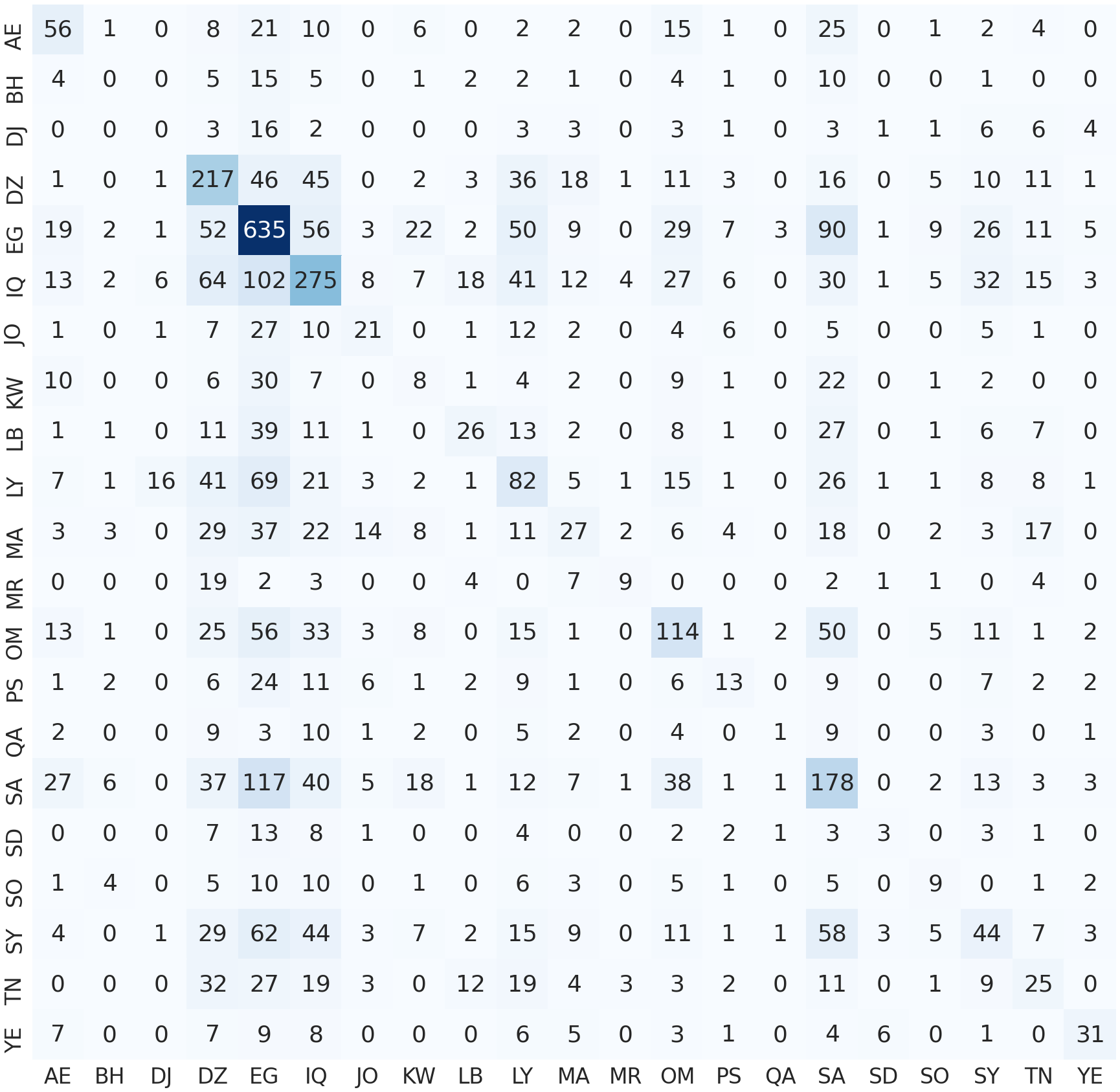}
         \caption{MSA identification}
         \label{fig:cm_msa}
     \end{subfigure}
     \hfill
     \begin{subfigure}[b]{0.49\textwidth}
         \centering
         \includegraphics[scale=0.130]{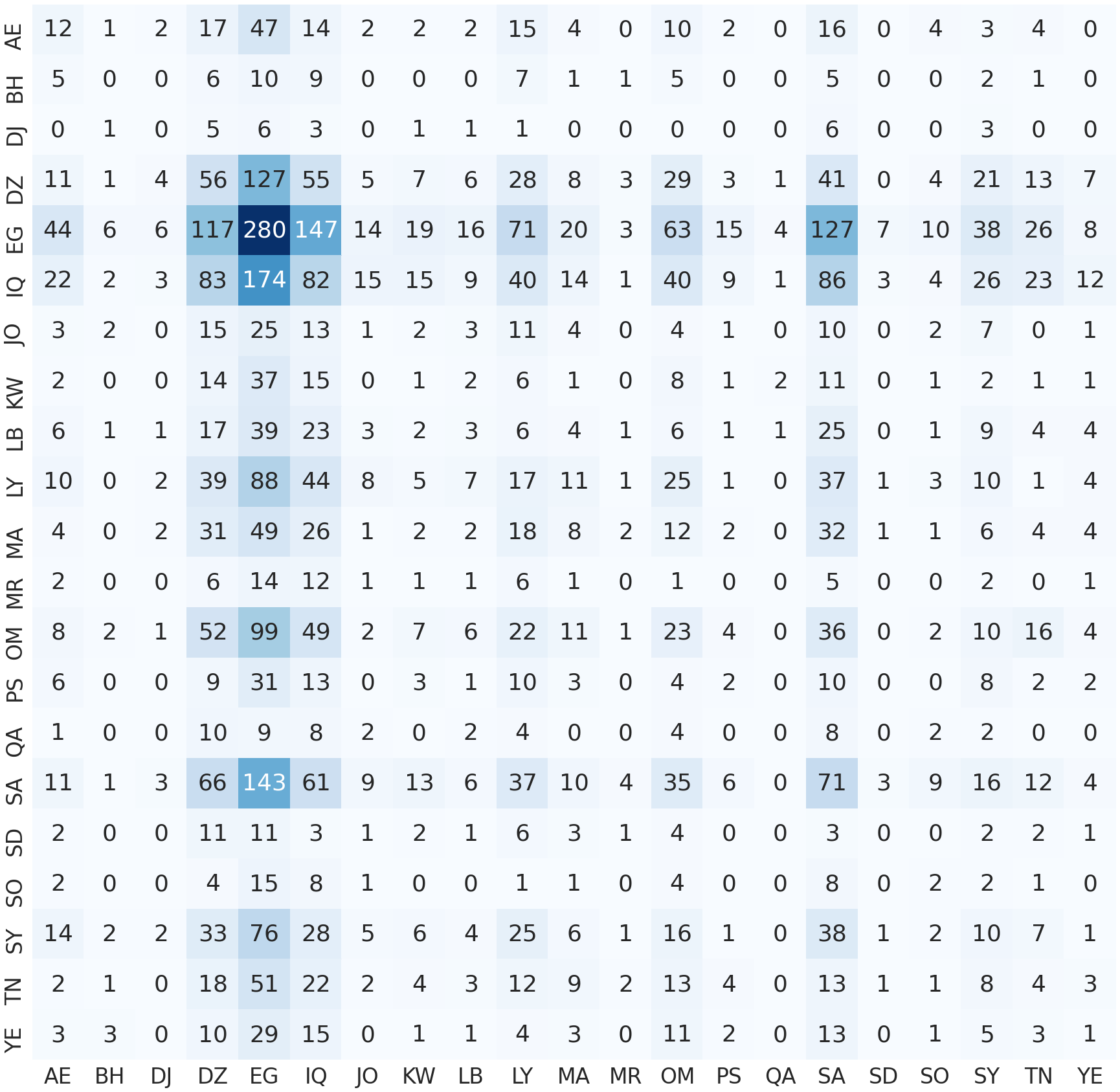}
         \caption{DA identification}
         \label{fig:cm_da}
     \end{subfigure}
        \caption{The confusion matrices of our MTL model on the Dev set for country-level DA and MSA identification.}
        \label{fig:cm}
\end{figure*}

\section{Results} \label{sec:results}
In our experiments, we have investigated multiple models, starting from traditional machine learning techniques to transformer-based approaches. The obtained results show that MARBERT significantly outperforms the other approaches. 
For a fair comparison, our single-task model employs an attention layer over the contextualized word embedding of MARBERT and concatenates its outputs with the [CLS] token embedding for MSA and DA identification subtasks. It is worth mentioning that incorporating an attention layer into single-task and MTL models improves the results compared to performing the classification using only the [CLS] token embedding representation.

Table \ref{table:results} presents the Macro-averaged F1-scores and the accuracies achieved using our evaluated single-task and MTL models on the four subtasks. The obtained results show that our attention-based multi-task model largely outperforms the single-task model on all subtasks' Dev set and Test set. For MSA identification, at the country-level and province-level,  our MTL model achieves F1-score performance increments of 1.49\% and 0.35\% respectively over the single-task model on the Dev set, while it achieves F1-score performance increments of 0.47\% and 0.63\% respectively over the single-task model for the Test set. For DA identification, at the country-level and province-level, the MTL model leads to F1-score performance increments of 0.42\% and 0.42\% over the single-task model, respectively on the Dev set, while it achieves increments of 1.57\% and 2.02\%, respectively on the Test set. This can be explained by the fact that our MTL model leverages signals from related tasks and boosts the performance of both. Moreover, through the task-specific attention layers, the MTL model extracts the task-discriminative features. Furthermore, employing task-specific features and global-shared features ([CLS] embedding) improves the performance of our MTL model.

\section{Discussion} \label{sec:discussion}
In order to analyze the results and explain the lower performance of our MTL model on some subtasks, Figure \ref{fig:cm} draws the confusion matrices for DA and MSA identification at the country-level on the Dev set.  The matrices highlight a number of strengths and weaknesses of our final model. On the one hand, the model performs well for countries with a high number of training examples such as Egypt, Algeria, and Iraq. On the other hand, the model shows poor performance in predicting true positives for countries with a low number of training examples, as it is the case of Djibouti and Bahrain. Moreover, the MSA country-level identification is a very challenging subtask since it is hard to find patterns to discriminate between countries and provinces based on standard language. Also, the model tends to make incorrect predictions for countries that are geographically close since their dialects have some minor differences (e.g. the Gulf countries) compared to other Arab countries.

\section{Conclusion} \label{sec:conclusion}
In this paper, we introduced our submitted system to the second NADI shared task. We proposed an MTL model for joint country-level and province-level identification of MSA and DA tweets. The model is based on the state-of-the-art MARBERT encoder and uses two task-specific attention layers to extract the task-discriminative features. The obtained results have shown that our MTL model outperforms the single-task model on all subtasks for both evaluation measures (Macro-F1 and accuracy).  

Future research work will focus on developing task-interaction and class-interaction modules and mechanisms for coarse-grained and fine-grained DA and MSA identification.

\bibliography{anthology,eacl2021}

\begin{thebibliography}{18}
\expandafter\ifx\csname natexlab\endcsname\relax\def\natexlab#1{#1}\fi

\bibitem[{Abdul-Mageed et~al.(2020{\natexlab{a}})Abdul-Mageed, Elmadany, and
  Nagoudi}]{mageed2020marbert}
Muhammad Abdul-Mageed, AbdelRahim Elmadany, and El~Moatez~Billah Nagoudi.
  2020{\natexlab{a}}.
\newblock {ARBERT} \& {MARBERT}: Deep bidirectional transformers for arabic.
\newblock \emph{arXiv preprint arXiv:2101.01785}.

\bibitem[{Abdul-Mageed et~al.(2020{\natexlab{b}})Abdul-Mageed, Zhang, Bouamor,
  and Habash}]{mageed:2020:nadi}
Muhammad Abdul-Mageed, Chiyu Zhang, Houda Bouamor, and Nizar Habash.
  2020{\natexlab{b}}.
\newblock {NADI 2020: The First Nuanced Arabic Dialect Identification Shared
  Task}.
\newblock In \emph{Proceedings of the Fifth Arabic Natural Language Processing
  Workshop (WANLP 2020)}, Barcelona, Spain.

\bibitem[{Abdul-Mageed et~al.(2021)Abdul-Mageed, Zhang, Elmadany, Bouamor, and
  Habash}]{mageed:2021:nadi}
Muhammad Abdul-Mageed, Chiyu Zhang, AbdelRahim Elmadany, Houda Bouamor, and
  Nizar Habash. 2021.
\newblock {NADI 2021: The Second Nuanced Arabic Dialect Identification Shared
  Task}.
\newblock In \emph{Proceedings of the Sixth Arabic Natural Language Processing
  Workshop (WANLP 2021)}.

\bibitem[{Abdul-Mageed et~al.(2020{\natexlab{c}})Abdul-Mageed, Zhang, Elmadany,
  and Ungar}]{abdul-mageed-etal-2020-toward}
Muhammad Abdul-Mageed, Chiyu Zhang, AbdelRahim Elmadany, and Lyle Ungar.
  2020{\natexlab{c}}.
\newblock \href {https://doi.org/10.18653/v1/2020.emnlp-main.472} {Toward
  micro-dialect identification in diaglossic and code-switched environments}.
\newblock In \emph{Proceedings of the 2020 Conference on Empirical Methods in
  Natural Language Processing (EMNLP)}, pages 5855--5876, Online. Association
  for Computational Linguistics.

\bibitem[{Abu~Kwaik and Saad(2019)}]{abu-kwaik-saad-2019-arbdialectid}
Kathrein Abu~Kwaik and Motaz Saad. 2019.
\newblock \href {https://doi.org/10.18653/v1/W19-4632} {{A}rb{D}ialect{ID} at
  {MADAR} shared task 1: Language modelling and ensemble learning for fine
  grained {A}rabic dialect identification}.
\newblock In \emph{Proceedings of the Fourth Arabic Natural Language Processing
  Workshop}, pages 254--258, Florence, Italy. Association for Computational
  Linguistics.

\bibitem[{Al-Sughaiyer and Al-Kharashi(2004)}]{msa_morphological_complexity}
Imad~A. Al-Sughaiyer and Ibrahim~A. Al-Kharashi. 2004.
\newblock \href {https://doi.org/https://doi.org/10.1002/asi.10368} {Arabic
  morphological analysis techniques: A comprehensive survey}.
\newblock \emph{Journal of the American Society for Information Science and
  Technology}, 55(3):189--213.

\bibitem[{Bahdanau et~al.(2015)Bahdanau, Cho, and
  Bengio}]{Bahdanau2015NeuralMT}
Dzmitry Bahdanau, Kyunghyun Cho, and Yoshua Bengio. 2015.
\newblock Neural machine translation by jointly learning to align and
  translate.
\newblock \emph{CoRR}, abs/1409.0473.

\bibitem[{Bouamor et~al.(2019)Bouamor, Hassan, and Habash}]{bouamor:2019:madar}
Houda Bouamor, Sabit Hassan, and Nizar Habash. 2019.
\newblock {The MADAR shared task on Arabic fine{-}grained dialect
  identification}.
\newblock In \emph{Proceedings of the Fourth Arabic Natural Language Processing
  Workshop}, pages 199--207.

\bibitem[{Devlin et~al.(2019)Devlin, Chang, Lee, and
  Toutanova}]{devlin-etal-2019-bert}
Jacob Devlin, Ming-Wei Chang, Kenton Lee, and Kristina Toutanova. 2019.
\newblock \href {https://doi.org/10.18653/v1/N19-1423} {{BERT}: Pre-training of
  deep bidirectional transformers for language understanding}.
\newblock In \emph{Proceedings of the 2019 Conference of the North {A}merican
  Chapter of the Association for Computational Linguistics: Human Language
  Technologies, Volume 1 (Long and Short Papers)}, pages 4171--4186,
  Minneapolis, Minnesota. Association for Computational Linguistics.

\bibitem[{El~Mekki et~al.(2020)El~Mekki, Alami, Alami, Khoumsi, and
  Berrada}]{el-mekki-etal-2020-weighted}
Abdellah El~Mekki, Ahmed Alami, Hamza Alami, Ahmed Khoumsi, and Ismail Berrada.
  2020.
\newblock \href {https://www.aclweb.org/anthology/2020.wanlp-1.27} {Weighted
  combination of {BERT} and n-{GRAM} features for nuanced {A}rabic dialect
  identification}.
\newblock In \emph{Proceedings of the Fifth Arabic Natural Language Processing
  Workshop}, pages 268--274, Barcelona, Spain (Online). Association for
  Computational Linguistics.

\bibitem[{Habash et~al.(2012)Habash, Diab, and
  Rambow}]{habash-etal-2012-conventional}
Nizar Habash, Mona Diab, and Owen Rambow. 2012.
\newblock \href
  {http://www.lrec-conf.org/proceedings/lrec2012/pdf/579_Paper.pdf}
  {Conventional orthography for dialectal {A}rabic}.
\newblock In \emph{Proceedings of the Eighth International Conference on
  Language Resources and Evaluation ({LREC}-2012)}, pages 711--718, Istanbul,
  Turkey. European Languages Resources Association (ELRA).

\bibitem[{Habash(2010)}]{habash2010introduction}
Nizar~Y Habash. 2010.
\newblock \emph{Introduction to Arabic natural language processing}, volume~3.
\newblock Morgan \& Claypool Publishers.

\bibitem[{Kingma and Ba(2014)}]{kingma2017adam}
Diederik~P. Kingma and Jimmy Ba. 2014.
\newblock \href {http://arxiv.org/abs/1412.6980} {Adam: A method for stochastic
  optimization}.
\newblock Cite arxiv:1412.6980Comment: Published as a conference paper at the
  3rd International Conference for Learning Representations, San Diego, 2015.

\bibitem[{Meftouh et~al.(2019)Meftouh, Abidi, Harrat, and
  Smaili}]{meftouh-etal-2019-smart}
Karima Meftouh, Karima Abidi, Salima Harrat, and Kamel Smaili. 2019.
\newblock \href {https://doi.org/10.18653/v1/W19-4633} {The {SM}ar{T}
  classifier for {A}rabic fine-grained dialect identification}.
\newblock In \emph{Proceedings of the Fourth Arabic Natural Language Processing
  Workshop}, pages 259--263, Florence, Italy. Association for Computational
  Linguistics.

\bibitem[{Salameh et~al.(2018)Salameh, Bouamor, and
  Habash}]{salameh-etal-2018-fine}
Mohammad Salameh, Houda Bouamor, and Nizar Habash. 2018.
\newblock \href {https://www.aclweb.org/anthology/C18-1113} {Fine-grained
  {A}rabic dialect identification}.
\newblock In \emph{Proceedings of the 27th International Conference on
  Computational Linguistics}, pages 1332--1344, Santa Fe, New Mexico, USA.
  Association for Computational Linguistics.

\bibitem[{Talafha et~al.(2020)Talafha, Ali, Za{'}ter, Seelawi, Tuffaha, Samir,
  Farhan, and Al-Natsheh}]{talafha-etal-2020-multi}
Bashar Talafha, Mohammad Ali, Muhy~Eddin Za{'}ter, Haitham Seelawi, Ibraheem
  Tuffaha, Mostafa Samir, Wael Farhan, and Hussein Al-Natsheh. 2020.
\newblock \href {https://www.aclweb.org/anthology/2020.wanlp-1.10}
  {Multi-dialect {A}rabic {BERT} for country-level dialect identification}.
\newblock In \emph{Proceedings of the Fifth Arabic Natural Language Processing
  Workshop}, pages 111--118, Barcelona, Spain (Online). Association for
  Computational Linguistics.

\bibitem[{Yang et~al.(2016)Yang, Yang, Dyer, He, Smola, and
  Hovy}]{yang-etal-2016-hierarchical}
Zichao Yang, Diyi Yang, Chris Dyer, Xiaodong He, Alex Smola, and Eduard Hovy.
  2016.
\newblock \href {https://doi.org/10.18653/v1/N16-1174} {Hierarchical attention
  networks for document classification}.
\newblock In \emph{Proceedings of the 2016 Conference of the North {A}merican
  Chapter of the Association for Computational Linguistics: Human Language
  Technologies}, pages 1480--1489, San Diego, California. Association for
  Computational Linguistics.

\bibitem[{Zaidan and Callison-Burch(2014)}]{zaidan_paper_taxonomy}
Omar~F. Zaidan and Chris Callison-Burch. 2014.
\newblock \href {https://doi.org/10.1162/COLI\_a\_00169} {Arabic dialect
  identification}.
\newblock \emph{Computational Linguistics}, 40(1):171--202.

\end{thebibliography}
\bibliographystyle{acl_natbib}

\end{document}